\documentclass[journal]{IEEEtran}
\IEEEoverridecommandlockouts
\usepackage{graphicx}%
\usepackage{multirow}%
\usepackage{amsmath,amssymb,amsfonts}%
\usepackage{amsthm}%
\usepackage{mathrsfs}%
\usepackage{xcolor}%
\usepackage{textcomp}%
\usepackage{manyfoot}%
\usepackage{booktabs}%
\usepackage{algorithm}%
\usepackage{algorithmicx}%
\usepackage{algpseudocode}%
\usepackage{listings}%
\usepackage{url}
\theoremstyle{thmstyleone}%
\theoremstyle{thmstyletwo}%

\theoremstyle{thmstylethree}%

\begin{document}

\title{Converging Paradigms: The Synergy of Symbolic and Connectionist AI in LLM-Empowered Autonomous Agents
}

\author{Haoyi Xiong, Zhiyuan Wang, Xuhong Li, Jiang Bian, Zeke Xie, Shahid Mumtaz, Anwer Al-Dulaimi, and Laura E. Barnes\thanks{H. Xiong and J. Bian are with Microsoft Corporation. Z. Wang and L. E. Barnes are with the University of Virginia. Xuhong Li is with Univiersit{\'e} de technologie de Compi{\`e}gne. Z. Xie is with Hong Kong University of Science and Technology (GZ). S. Mumtaz is with Nottingham Trent University, Nottingham, UK and Silesian University of Technology, Gliwice, Poland. Anwer Al-Dulaimi is with College of Technological Innovation, Zayed University, Abu Dhabi, UAE. }}


\maketitle

\begin{abstract}
   This article explores the convergence of connectionist and symbolic artificial intelligence (AI), from historical debates to contemporary advancements. Traditionally considered distinct paradigms, connectionist AI focuses on neural networks, while symbolic AI emphasizes symbolic representation and logic. Recent advancements in large language models (LLMs), exemplified by ChatGPT and GPT-4, highlight the potential of connectionist architectures in handling human language as a form of symbols. The study argues that LLM-empowered Autonomous Agents (LAAs) embody this paradigm convergence. By utilizing LLMs for text-based knowledge modeling and representation, LAAs integrate neuro-symbolic AI principles, showcasing enhanced reasoning and decision-making capabilities. Comparing LAAs with Knowledge Graphs within the neuro-symbolic AI theme highlights the unique strengths of LAAs in mimicking human-like reasoning processes, scaling effectively with large datasets, and leveraging in-context samples without explicit re-training. The research underscores promising avenues in neuro-vector-symbolic integration, instructional encoding, and implicit reasoning, aimed at further enhancing LAA capabilities. By exploring the progression of neuro-symbolic AI and proposing future research trajectories, this work advances the understanding and development of AI technologies. 
\end{abstract}
%


\section{Introduction}\label{sec1}
Artificial Intelligence (AI) has historically navigated the fascinating duality of two foundational paradigms: connectionism and symbolism. Connectionism, deeply influenced by cognitive science and computational neuroscience, delves into neural networks and machine learning algorithms that echo the deep neural architecture and functions of the human brain~\cite{lecun2015deep}. Imagine a sprawling network of neurons firing in electric synchrony, mirroring how advanced AI systems identify patterns and glean insights from vast datasets. Conversely, symbolism is the epitome of conceptual and logical clarity. It anchors itself in the high-level abstractions and representations of knowledge, flourishing through rule-based systems that excel in reasoning and decision-making~\cite{newell2007computer}. Picture a grand library where every book is a rule, and every chapter a pathway to logical deduction--symbolic AI analogising the thought processes of human reasoning.

The dynamic interplay between these two paradigms has sculpted the continuous evolution of AI, like a grand philosophical debate, resulting in shifts in dominance and application across various research domains. Think of this dialectic as a dance through time—the elegant waltz of connectionism and symbolism, sometimes leading, sometimes following, yet always in a harmonious exchange that propels the boundaries of what AI can achieve. For instance, in the domain of image recognition, connectionist models driven by deep neural networks demonstrate their prowess by identifying subtle patterns in pixel data, akin to how our brains recognize faces in a crowd~\cite{krizhevsky2012imagenet}. Meanwhile, in expert systems used for medical diagnostics, symbolism shines by methodically applying predefined rules to diagnose diseases, mimicking the logical flow of a doctor's thought process~\cite{shortliffe2012computer}. This storied dance of paradigms has not just shaped, but revitalized AI, continuing to impact its trajectory as it ventures into increasingly sophisticated applications. The oscillation of dominance between these approaches resembles the ebb and flow of tides, each rise and retreat bringing new insights and innovations to the fore.  

\begin{figure*}
    \centering
    \includegraphics[width=0.85\textwidth]{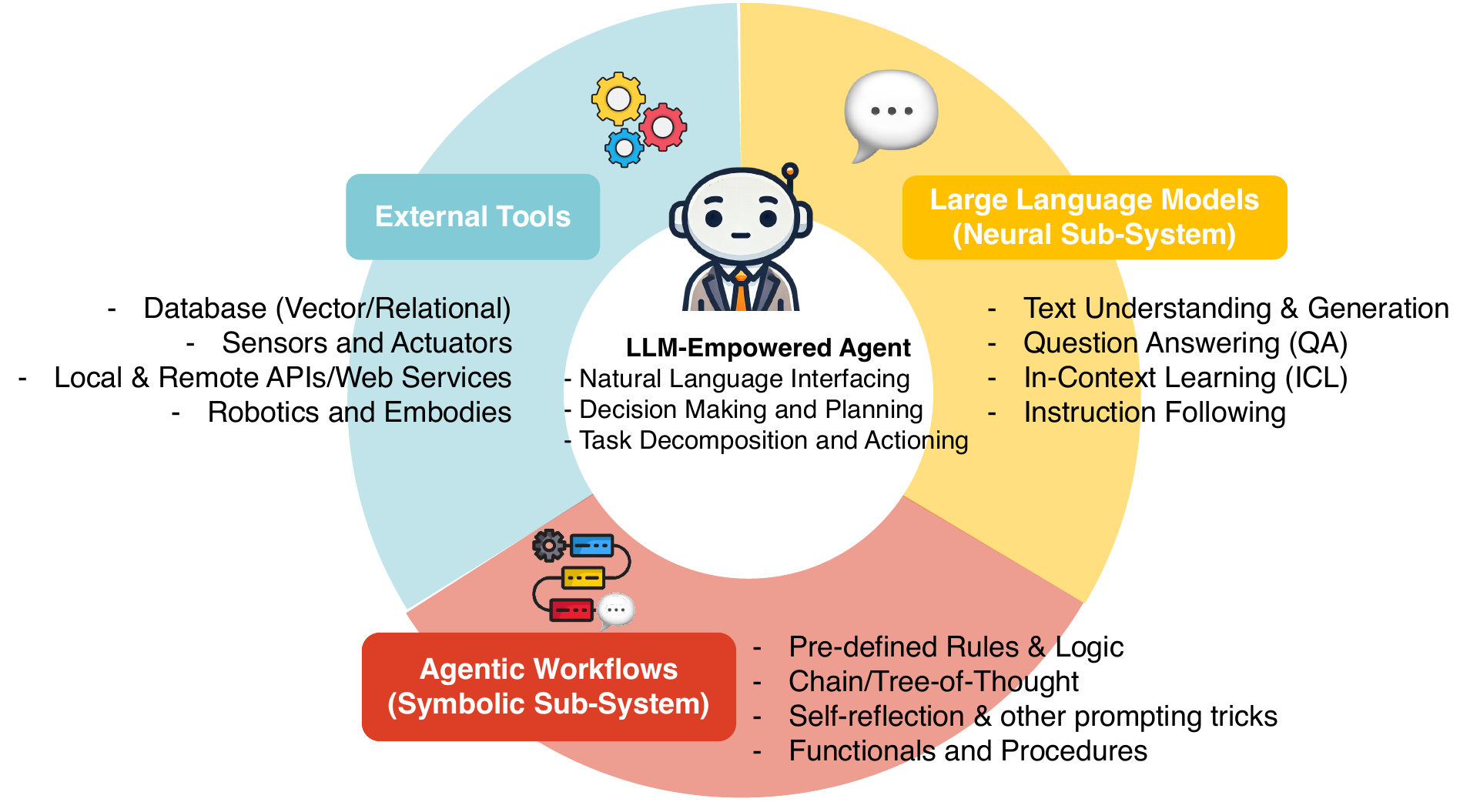}
    \caption{Elements of LLM-empowered Autonomous Agents (LAAs): Large Language Models (Neural Sub-System), Agentic Workflows (Symbolic Sub-System), and External Tools}
    \label{fig:intro-LAA}
\end{figure*}

In recent years, the advancements in Large Language Models (LLMs) and foundation models have catalyzed the integration of connectionist and symbolic AI paradigms, realizing new levels of computational intelligence and versatility~\cite{brown2020language}. These models, exemplified by systems such as OpenAI's GPT-4, have demonstrated unprecedented capabilities in natural language understanding and generation, exhibiting robust performance across a range of complex tasks~\cite{radford2019language}. LLMs themselves are a triumph of connectionism, empowered by vast amounts of data and sophisticated neural architectures to produce coherent and contextually relevant texts. Moreover, the emergence of LLM-empowered Autonomous Agents (LAAs) signifies a pivotal juncture in the development of AI, embodying the convergence of symbolic and connectionist AI. As shown in Figure~\ref{fig:intro-LAA}, LAAs combine a symbolic subsystem, utilizing language-based knowledge, rules, and workflows intrinsic to symbolic AI, with the generative capabilities of LLMs~\cite{bubeck2023sparks}. This symbolic subsystem works seamlessly with the neural subsystem and incorporates external tools for perceptions and actions~\cite{xiong2023natural}. LAAs demonstrate advanced reasoning, planning, and decision-making abilities, marking a new era in AI. The dual subsystems align with dual-process theories of reasoning~\cite{evans2003two} and Systems I and II proposed by Yoshua Bengio~\cite{bengio2020deep}.  



In this work, we aim at examining the historical evolution and current state of AI by exploring the enduring debate between connectionism and symbolism and their convergence in modern technologies, particularly in the theme of neuro-symbolic approaches, including Knowledge Graphs, LLMs, and LAAs. This review aims to illustrate how the integration of these paradigms has led to groundbreaking advancements, offering new perspectives on the capabilities and future directions of AI.
\begin{itemize}

    \item \textbf{Historical Context of Technology}: This article provides an in-depth examination of the historical debate between connectionism and symbolism, contextualizing modern AI developments and highlighting the strengths of each approach. We present recent advancements in LLMs with Knowledge Graphs (KGs)~\cite{ji2021survey} as references, discussing these techniques from the perspectives of symbolic, connectionist, and neuro-symbolic AI. The article also showcases the transformative impact of these techniques on knowledge modeling, acquisition, representation, and reasoning.

    \item \textbf{Convergence of Paradigms}: This article highlights the convergence of symbolic and connectionist approaches in developing LAAs, emphasizing their enhanced reasoning, decision-making, and efficiency. By contrasting LAAs with Knowledge Graphs (KGs) within neuro-symbolic AI, we examine distinct patterns and functionalities. While both integrate symbolic and neural methodologies, LAAs demonstrate unique advantages over KGs: (1) analogizing human reasoning with agentic workflows and various prompting techniques~\cite{webb2023emergent,strachan2024testing}, (2) scaling effectively on large datasets, adapting to in-context samples, and leveraging the emergent abilities of LLMs. These strengths drive the surge of a new wave of neuro-symbolic AI~\cite{garcez2023neurosymbolic}.
    
    
    \item \textbf{Future Directions}: The article highlights the trend of converging paradigms and current limitations of LAAs, pointing to two promising future directions. First is the development of \emph{neuro-vector-symbolic architectures}, which integrate vector manipulation to enhance the reasoning capabilities of agents. Second is the approach known as \emph{program-proof-of-thoughts (P$^2$oT) prompting}. This involves breaking down complex reasoning processes into verifiable propositions, utilizing program proof languages (such as Dafny) for structured verification. It aims to provide rigorous reasoning by modeling propositions, integrating with theorem provers, and focusing on applications in specific domains.
    
    
    
\end{itemize}
These contributions are crucial as they provide a comprehensive understanding of the evolution of AI, highlight the significance of paradigm convergence, and offer insights into future research and application potentials in the rapidly evolving field of AI.  

\section{Preliminaries}
This section begins by summarizing the historical debate between connectionist AI and symbolic AI. We then explore knowledge graphs (KGs) as an early effort to synergize these two paradigms through neuro-symbolic AI. Lastly, we examine LLMs as the latest advancements in connectionist AI.  

\begin{figure*}
    \centering
    \includegraphics[width=0.98\textwidth]{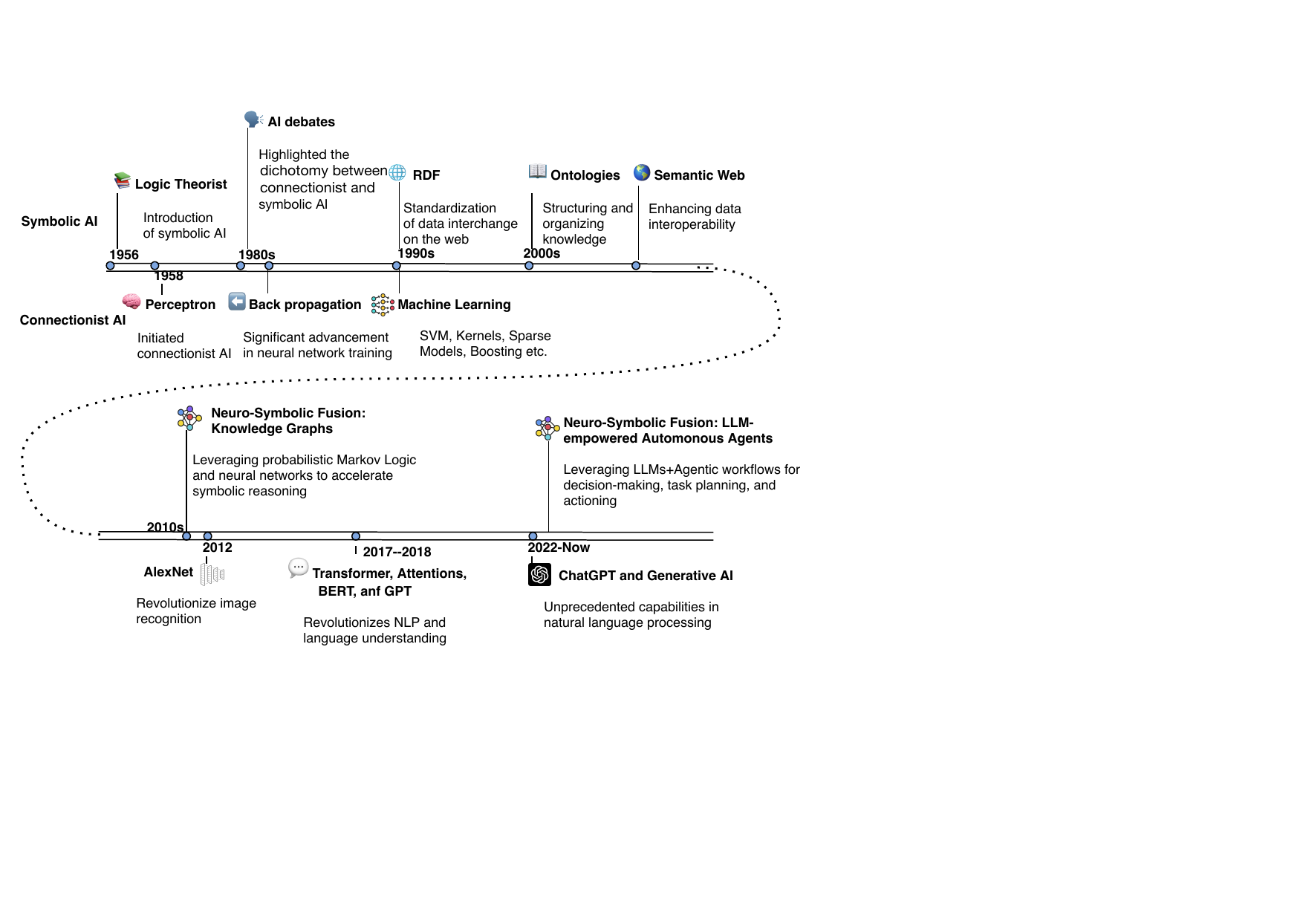}
    \caption{\textbf{Exploring the Evolution of Artificial Intelligence: A Timeline of Key Innovations and Milestones}. It starts from the birth of symbolic and connectionist AI in the 1950s, through key milestones like the AI debates of the 1980s and the advancement in machine learning in the 1990s. This figure highlights significant developments such as the impact of AlexNet on image recognition, the transformation in NLP by models like BERT and GPT, and the rise of generative AI, culminating in the use of LLMs and Agents for autonomous decision-making in the 2020s.}
    \label{fig:ai-evolution}
\end{figure*}

\subsection{Connectionism vs. Symbolism: a Historical Debate on AI}
As shown in Figure~\ref{fig:ai-evolution}, the discourse of AI has long revolved around the dichotomy between connectionism and symbolism, two paradigms integral to the field. Connectionism models cognitive processes through artificial neural networks that emulate the brain's neuron structures, emphasizing learning through algorithms and pattern recognition. This began with Frank Rosenblatt's Perceptron in 1958~\cite{rosenblatt1958perceptron} and advanced significantly with the backpropagation algorithm developed by David Rumelhart, Geoffrey Hinton, and Ronald J. Williams in the 1980s~\cite{rumelhart1986learning}, setting the stage for modern deep learning~\cite{lecun2015deep}.
Conversely, symbolism focuses on high-level knowledge representations and symbolic manipulation to mimic human reasoning, gaining prominence with systems like the Logic Theorist by Allen Newell and Herbert A. Simon in 1956~\cite{newell1956logic}. Symbolic AI thrived with expert systems such as MYCIN~\cite{shortliffe2012computer} and DENDRAL~\cite{buchanan1978dendral} in the 1970s and 1980s, excelling in specific domains through predefined rules.

In the 1980s, as Ashok Goel noted, debates often involved criticisms that attacked caricatures of the opposing methods~\cite{goel1989integration}. Each approach has its limitations: connectionist AI is criticized for its black-box nature and lack of interpretability~\cite{lipton2018mythos}, while symbolic AI faced challenges with the labor-intensive knowledge acquisition process~\cite{feigenbaum1977art} and its limited adaptability~\cite{elkan1993building}. 
Historical debates between figures, such as Yann LeCun, Yoshua Bengio, and Gary Marcus, have underscored these limitations~\cite{bengio2021deep}. However, the integration of both paradigms has led to robust hybrid models, combining neural networks' pattern recognition with symbolic systems' interpretability and logical reasoning~\cite{garcez2008neural}. Contemporary research exemplifies this convergence, seen in neuro-symbolic AI and large-scale pre-trained models like BERT~\cite{devlin2018bert}, GPT~\cite{brown2020language}, and hybrid reinforcement learning models~\cite{silver2017mastering}, reflecting the ongoing evolution inspired by the historical debate.


\subsection{Knowledge Graphs: An Early Neuro-symbolic Attempt}

Knowledge graphs have a foundation rooted in the evolution of semantic web technologies and the Resource Description Framework (RDF). Proposed by the W3C in the 1990s, RDF standardized data interchange on the web using triples (subject, predicate, object) for seamless data integration and interoperability~\cite{W3C1999}. This movement established the Semantic Web, aiming for a more intelligent and interconnected web~\cite{berners2023semantic}. Early adopters used RDF to build schemas and taxonomies, forming the basics of modern knowledge graphs~\cite{antoniou2004semantic}.


As the field matured, the focus shifted towards capturing complex relationships and domain-specific knowledge. Ontologies, formal specifications of concepts and relationships, provided a framework for annotating and interlinking data, enabling semantic reasoning at a certain level \cite{gruber1993translation}. Markov-logic networks introduced probabilistic reasoning to knowledge graphs, allowing for handling uncertainty and inconsistency in data~\cite{kok2005learning}. The synergy of Ontologies and Markov-logic networks advanced the ability of symbolic AI to perform robust reasoning over large datasets~\cite{nickel2015review}.

In recent years, the use of graph neural networks (GNNs) has further revolutionized the landscape of knowledge graphs. GNNs adeptly leverage the graph structure for advanced pattern recognition and complex predictions. They excel in tasks such as node classification, link prediction, and the extraction of hidden patterns from graph-structured data~\cite{kipf2016semi}. This paradigm shift towards neural networks marks a convergence with modern machine learning techniques, enabling more nuanced and scalable interpretations of often massive and intricate datasets. The ability of GNNs to embed nodes and entire graphs numerically has significantly enhanced the computational handling of knowledge graphs~\cite{wang2017knowledge}. In conclusion, the integration of graph neural networks with rule-based reasoning has positioned knowledge graphs at the core of the neuro-symbolic AI approach~\cite{ji2021survey} prior to the surge of LLMs.

\subsection{LLMs: Recent Connectionist AI Advancements}
The field of connectionist AI has undergone substantial evolution, beginning with the invention of the perceptron~\cite{rosenblatt1958perceptron}, kicking off the neural network research in the late 1950s. In the following decades, the development of Multi-Layer Perceptrons (MLPs) introduced hidden layers and non-linear activation functions, enabling the modeling of more complex functions~\cite{rumelhart1986learning}. In the 1990s, Long Short-Term Memory (LSTM) networks were developed to address the limitations of traditional recurrent neural networks (RNNs) by introducing gating mechanisms to handle long-term dependencies in sequential data~\cite{hochreiter1997long}. Self-attention mechanisms and transformer architectures proposed in the late 2010s further revolutionized sequence modeling, such as texts for natural language processing, by allowing models to focus on different parts of the input sequence when generating each part of the output sequence~\cite{vaswani2017attention}.

\begin{figure*}
    \centering
    \includegraphics[width=0.95\textwidth]{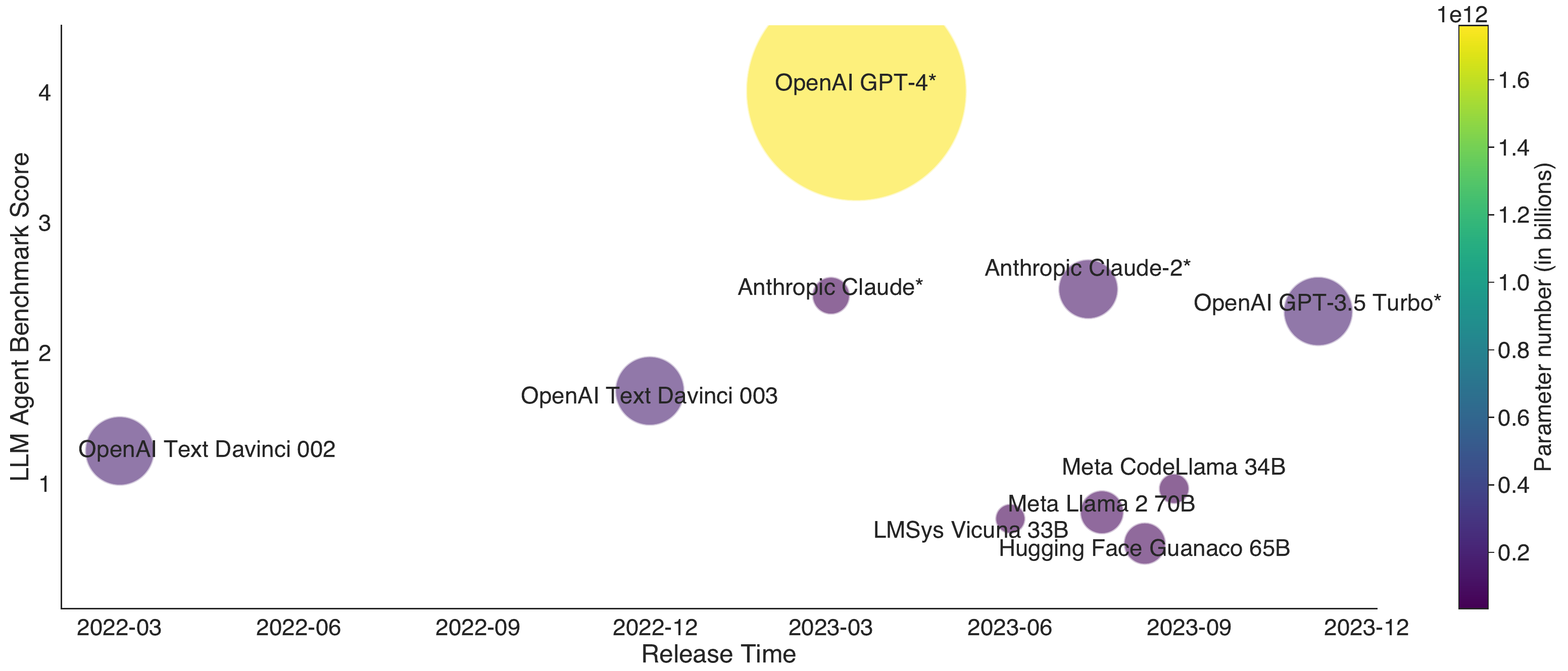}
    \caption{Large Language Models and Their Agentic Abilities. The X-axis shows the release dates, and the Y-axis represents the LLM Agent Benchmark Score~\cite{liu2023agentbench}. Bubble size indicates the number of parameters (in billions). An asterisk (*) denotes estimated parameter counts when the official release is not available.}
    \label{fig:llms}
\end{figure*}

The development of transformer-based pre-trained language models has significantly advanced natural language processing (NLP). These architectures include encoder-only models, e.g., BERT~\cite{devlin2018bert}, which excel at understanding and classifying text; decoder-only models, e.g., GPT~\cite{radford2019language}, which generate coherent and contextually relevant text; and encoder-decoder models, e.g., T5~\cite{raffel2020exploring}, which are effective in tasks requiring both comprehension and generation. Transformer-based language models, such as OpenAI's GPT-4~\cite{achiam2023gpt}, Google's Gemini~\cite{team2023gemini} and PaLM~\cite{chowdhery2023palm}, Microsoft's Phi-3~\cite{abdin2024phi}, and Meta's LLaMA~\cite{touvron2023llama}, are termed Large Language Models (LLMs). These models, illustrated in Figure~\ref{fig:llms}, are trained on large-scale transformers comprising billions of learnable parameters to support various abilities to enable agents, including perception, reasoning, planning, and action~\cite{webb2023emergent}. As the central component of an agent's neural sub-system, the larger the model, the stronger the agent's capability.


In general, every LLM undergos a two-stage training process: \emph{pre-training} and \emph{fine-tuning}. Pre-training involves adjusting model parameters based on the statistical properties of a large text corpus, enabling an understanding of syntax, semantics, and linguistic nuances~\cite{devlin2018bert}. Fine-tuning then adapts the pre-trained model to specific tasks or domains using a smaller, task-specific dataset, optimizing performance for particular applications~\cite{ding2023parameter}. To ensure LLMs follow human's instructions, align with human values and exhibit desired behaviors, instruction tuning and reinforcement learning from human feedback (RLHF) have been proposed on top of fine-tuning~\cite{ouyang2022training}.

As the size of LLMs increases, they exhibit a range of emerging capabilities, such as writing computer code, playing chess, diagnosing medical conditions, and translating languages. These capabilities often develop suddenly and dramatically at certain scales due to scaling laws, which describe how task performance can surge unexpectedly when a model reaches a particular threshold size~\cite{schaeffer2024emergent}. This phenomenon is particularly observable in tasks requiring multi-step reasoning, where success probabilities compound multiplicatively, leading to rapid performance jumps~\cite{wei2022chain}. However, these advancements come with \emph{``hallucination'' challenges}~\cite{li2023halueval}, such as producing false or nonsensical information that appears convincing but is inaccurate or not based on reality. These issues underline the importance of continued research and engineering to harness the benefits of LLMs while mitigating their drawbacks.

\section{LLM-empowered Autonomous Agents: The Convergence of Symbolism and Connectionism}
This section reviews the definition of both traditional and LLM-based agents, introduces core techniques for designing and implementing LAAs, and rethinks these innovations through the lens of symbolic AI.  

\subsection{Autonomous Agents: Classic and LLM-empowered}
An autonomous agent is an artificially intelligent entity designed to achieve specific goals independently, acquiring contextual factors to perceive the environmental state and undertaking context-relevant actions~\cite{maes1993modeling}. These agents, equipped with reasoning, learning, and adaptability, thrive in dynamic and complex contexts. Unlike traditional software programs that follow predetermined rules, autonomous agents operate with self-governing attributes, allowing them to function under varying conditions~\cite{albrecht2018autonomous}. Leveraging these capabilities, they facilitate automation by performing tasks that typically require human intervention, enhancing efficiency, and reducing operational costs across fields such as robotics, communication, financial trading, and healthcare \cite{albrecht2018autonomous}. For instance, in robotic applications, autonomous agents can navigate tasks with minimal supervision, continuously monitor their surroundings, and adapt to new situations, making them robust solutions for long-term automation \cite{arkin1998behavior,rizk2019cooperative}.

The foundational techniques of autonomous agent design originate from classic AI approaches, such as Probabilistic Graphical Models~\cite{koller2009probabilistic}, Reinforcement Learning~\cite{sutton2018reinforcement}, and Multi-Agent Systems~\cite{wang2024survey}, which manage uncertainty and learn optimal behaviors in dynamic environments or enable agents to interact and share information efficiently. However, the advent of LAAs marks a significant evolution beyond traditional AI for both symbolic and neural sub-systems. These agents use extensive pre-training on vast textual corpora to acquire broad knowledge, performing human reasoning tasks by generating contextually appropriate text \cite{xi2023rise}. This capability not only simulates understanding and decision-making but also allows the generation of code and other communicative texts, enhancing their practical utility \cite{wu2023autogen}. By integrating pre-trained language models with natural language understanding, LAAs adapt flexibly to diverse scenarios, expanding AI's potential in autonomous operations.

\subsection{Design and Implementation of LAAs}
Central to the design of an agent is its neural sub-system--an LLM, which functions as the core controller or coordinator. The LLM orchestrates with the agent's symbolic sub-system and external tools, including a planning and reasoning component for task decomposition and self-reflection, memory (both short-term and long-term), and a tool-use component that allows access to external information and functionalities. 

\begin{itemize}
    \item \textbf{Agentic  Workflow}: An agentic workflow combines planning, reasoning, memory management, tool integration, and user interfaces with LLMs. Frameworks, such as LangChain~\cite{LangChain2022} and LlamaIndex~\cite{Liu_LlamaIndex_2022}, help design these workflows.
    
    
    \item \textbf{Planner and Reasoner}: Advanced techniques such as chain-of-thought and tree-of-thought prompting~\cite{song2023llm} break down tasks into sub-tasks, with self-reflection allowing agents to critique and refine outputs~\cite{lin2024swiftsage}.
    
        
    \item \textbf{Memory Management}: Incorporates short-term memory for context and long-term memory using external storage, such as vector databases, enabling efficient information retrieval and enhanced reasoning~\cite{johnson2019billion,guu2020retrieval}.
    
    
    \item \textbf{Tool-Use \& Natural Language Interface (NLI) Integration}: Agents can access external tools, APIs, and models, deciding when and how to utilize them based on task goals~\cite{reedgeneralist,shen2024hugginggpt}. In addition, An effective NLI interprets user requests and communicates actions~\cite{gao2018neural}. Techniques, such as ReAct and MRKL, provide structured interaction steps (thought, action, action input, observation)~\cite{yao2022react,karpas2022mrkl}.

    
\end{itemize}
By integrating these components, LAAs can tackle complex tasks. However, challenges like limited context windows, long-term planning, and reliable interfaces remain, necessitating ongoing research and development.  


\subsection{Rethink LAAs from the Perspective of Neuro-symbolic AI}
Neuro-symbolic AI combines the strengths of neural networks and symbolic reasoning, producing decision-making processes that are both explicit and interpretable. In autonomous agents enhanced by LLMs, the latest advancements in deep neural networks are harnessed, while task decomposition and planning are guided by symbolic AI principles --- breaking complex tasks into discrete, logical steps that can be systematically analyzed and reasoned through \cite{garcez2015neural}. This fusion of symbolic structures and deep neural networks creates a powerful synergy, significantly boosting the capabilities of these agents.  

\paragraph{Symbolic Modeling and Neural Representation}
Classic symbolic AI represents knowledge using abstractions and symbols, utilizing explicit symbolic modeling such as rules and relationships to perform reasoning~\cite{brachman2004knowledge}. This approach typically involves well-defined logic and structured knowledge bases, enabling systems to behave based on pre-defined rules. In contrast, LAAs, driven by language models, represent knowledge in a more distributed and implicit manner. Instead of relying on explicit symbols and rules, these agents leverage vast amounts of corpus and self-supervised pre-training on language models to infer patterns and relationships from raw text~\cite{devlin2018bert}. The knowledge is embedded within the weights of LLMs, allowing for more flexible and context-driven reasoning. This advantage fundamentally contrasts with the rigidity of symbolic AI, providing LAAs with the ability to handle ambiguity and generate more human-like responses~\cite{brown2020language}.

\paragraph{Search-based Decision Making by Generation}
Given a complex goal requiring multiple steps to achieve, existing agent technologies either harness symbolic AI to systematically explore the space of potential actions or employ reinforcement learning to optimize the trajectory of these actions, efficiently partitioning complex tasks into manageable subtasks~\cite{sutton2018reinforcement}.  Within a LLM-empowered agent, the Chain-of-Thought (CoT) method guides LLMs to generate texts about intermediate reasoning steps, enhancing their cognitive task performance~\cite{wei2022chain}. By breaking tasks into logical sequences, CoT prompts encourage LLMs to structure their reasoning systematically. This method overcomes LLM limitations at the token level by enabling coherent, step-by-step elaboration of thought processes, improving problem-solving accuracy and reliability. More recently, Tree-of-Thought (ToT) prompting extends this approach by allowing LLMs to explore multiple reasoning paths simultaneously in a tree structure~\cite{yao2024tree} and the proposal of functional search over program generation, leveraging large language models (LLMs), successfully facilitates mathematical discoveries~\cite{romera2024mathematical}. These methods enhance LLM problem-solving abilities by promoting dynamic and reflective reasoning processes, closely mirroring symbolic reasoning techniques, on top of a neural basis.

\paragraph{Case-based Reasoning through In-context Learning}
An agent must adapt to new situations, while traditional methods rely on either re-training neural networks or deducing examples of new situations into rules for better reasoning. Within a LLM-empowered agent, few-shot in-context learning (ICL) has been proposed to utilize given examples into a prompt to generate appropriate responses that solve problems without explicit re-training the LLM~\cite{min2022rethinking}. This approach mimics the \emph{case-based reasoning}, a fundamental concept in symbolic AI, by leveraging explicit knowledge and experiences to tackle new problems. 
%
%
This enhances the model's ability to generalize from specific examples, effectively creating a neuro-symbolic mapping from presented examples to desired outcomes.

\paragraph{Neuro-symbolic Integration Driven by Emergent Abilities}
The emergent abilities of LLMs, such as contextual understanding, sequential reasoning, goal reformulation, and task decomposition, are surged by over-parameterized architectures and extensive pre-training corpora~\cite{schaeffer2024emergent}. Combining well-designed rules with the emergent abilities of LLMs enables agents to create and follow complex workflows, known as agentic workflows. By prompting large language models with instructions like ``let’s think step by step'', these models analogise human's reasoning processes and can exhibit logical and mathematical reasoning, thereby enhancing their structured reasoning skills~\cite{webb2023emergent,strachan2024testing}. This agentic approach allows LLMs to not only process but also proactively generate structured, logical, and adaptive reasoning pathways~\cite{xi2023rise}, significantly improving their problem-solving and decision-making capabilities, marking a pivotal evolution in neuro-symbolic AI technologies.

\section{Discussions and Future Directions}
In this section, we discuss the LLM-empowered autonomous agent by comparing it with an alternative neuro-symbolic approach—the Knowledge Graph—and then highlight future directions for this technology.  

\subsection{Comparative Analysis: LAAs versus KGs}
Previous sections have presented LAAs and KGs, both of which exemplify neuro-symbolic approaches to AI. We here compare these two methodologies to highlight the superior positioning of LAAs in the current wave of AI advancements.  

KGs harness the power of symbolic AI, organizing domain-specific knowledge through explicit relationships and rules. This design makes them highly effective in static environments where precision, interpretability, and predefined schemas are crucial. Their logical reasoning capabilities ensure that outputs are consistent and verifiable, which is paramount for applications needing clarity and exactitude in knowledge modeling~\cite{delong2023neurosymbolic}. In addition, the scalability of KGs is inherently limited by their requirements of explicit schema definitions and manual updates~\cite{peng2023knowledge}. As the volume of data grows, the complexity of managing and querying the graph increases significantly. The maintenance of a large-scale KG demands substantial computational resources and human expertise, affecting efficiency and agility in evolving environments.

On the other hand, LAAs are designed with a more dynamic and flexible approach. By combining the language comprehension and generation abilities of neural networks with the structured reasoning of symbolic AI, these agents are equipped to tackle a wide range of complex tasks. The implicit knowledge stored in neural networks enables context-sensitive responses and seamless adaptation to changing environments~\cite{xie2021explanation}. Additionally, LLMs efficiently compress vast corpora into a learnable network, making these agents highly scalable. Once trained, the models can be fine-tuned with additional data at a fraction of the cost and effort required for updating knowledge graphs, and can even support in-context learning without fine-tuning. As a result, LLM-powered agents can handle larger datasets with ease and even process online data to respond to real-time changes effectively.  

Furthermore, the advanced reasoning mechanisms employed by LAAs, such as CoT~\cite{wei2022chain} and ToT~\cite{yao2024tree}, enable them to break down and solve complex problems effectively through analogising human reasoning steps~\cite{webb2023emergent}. These methods mitigate the limitations of token-level constraints in LLMs, fostering a more robust and contextually aware decision-making process. As a result, LAAs are poised to drive future innovations in AI, offering more versatile and intelligent solutions than their knowledge graph counterparts.

\subsection{Future Directions}
Following prior discussions, we propose several future research directions aimed at enhancing the current landscape of LAAs.

\begin{figure*}
    \centering
    \includegraphics[width=0.85\textwidth]{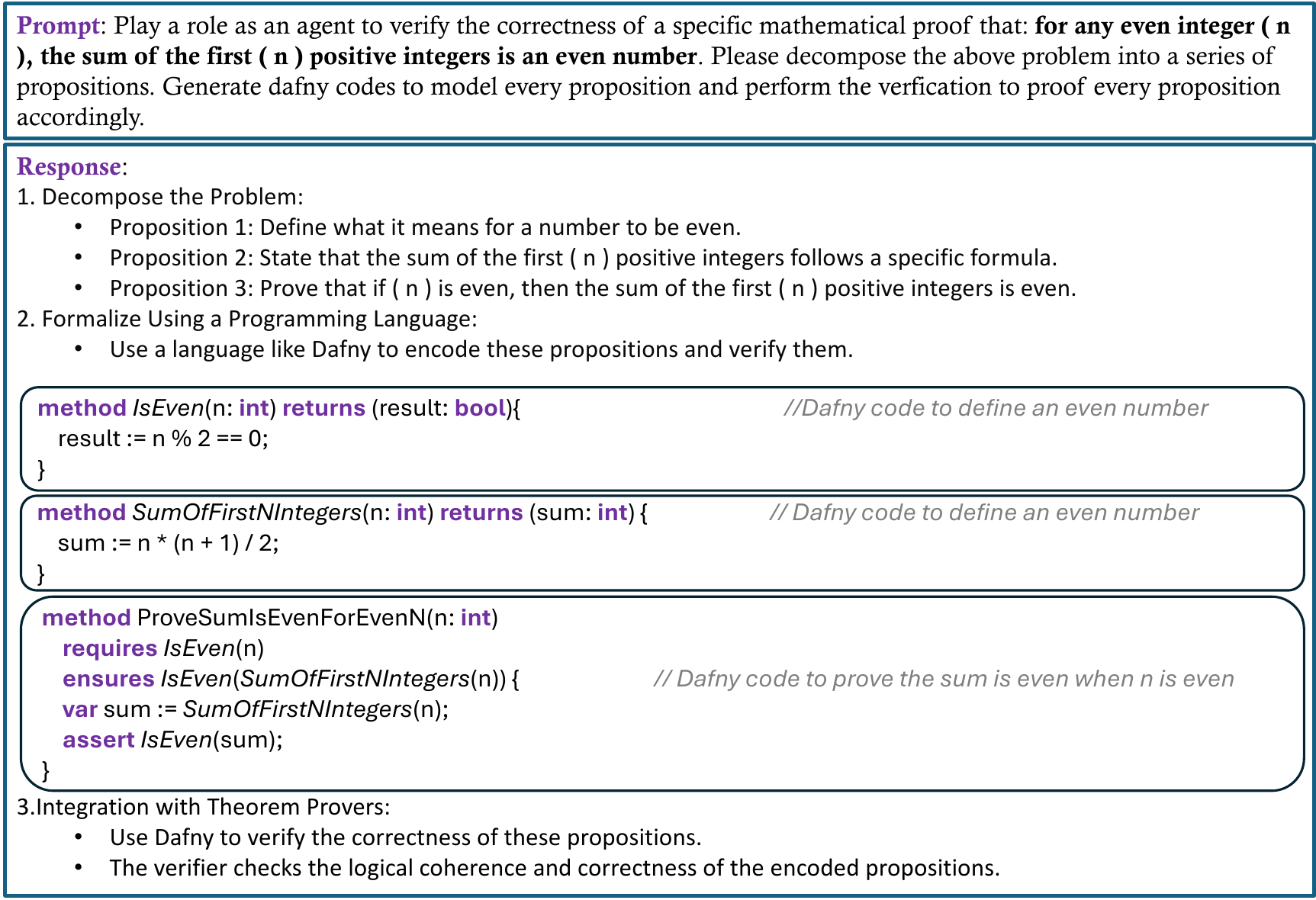}
    \caption{An Illustrative Example of Program-Proof-of-Thoughts (P$^2$oT) for Mathematical Proof Verification}
    \label{fig:pot-exp}
\end{figure*}

\subsection{Neuro-vector-symbolic Integrative Intelligence}
Current agentic reasoning approaches emulate human reasoning steps explicitly \cite{webb2023emergent}. For instance, when an agent receives a user's request, it retrieves similar cases and enhances its actions through in-context learning, and for ambiguous requests, the agent prompts the LLM to clarify and rewrite the request in various forms \cite{askari2024self}. This process involve extracting vectors for each rewritten request and performing multi-vector retrieval, improving context understanding and generative performance but increasing computational load. A vector-centric perspective, utilizing encoder-decoder architectures such as GritLM \cite{muennighoff2024generative} that prompt generative models for instructional text encoding/vectorization, implicit neural reasoners that extend transformers with causal relation graphs for enhanced long-range reasoning \cite{velivckovic2021neural} with latent vectors and attention  matrices, and vector-symbolic architectures (VSAs)~\cite{kanerva2009hyperdimensional}, could significantly address this problem.  Specifically, the VSA employs high-dimensional vectors to encode and manipulate information, allowing the representation of complex structures and relationships compactly and contextually~\cite{kanerva2009hyperdimensional}. It models the cognitive and reasoning processes as \emph{algebraic operations} in the vector space. Combining VSAs with LLMs could enhance cognitive capabilities, enabling precise multi-step decision-making, with applications in scientific discovery, such as solving Raven’s progressive matrices \cite{hersche2023neuro}, thus accelerating the convergence between connectionist and symbolic paradigms through computable vectorization.

\subsection{Program-Proof-of-Thoughts Reasoning}
We here illustrate the proposal of Program-Proof-of-Thoughts (P$^2$oT) for agentic reasoning in a rigorous manner,  building on the methodologies of CoT and ToT prompting. Specifically, P$^2$oT decomposes complex reasoning processes into a series of propositions organized in linear or tree structures. It leverages the programming language for program proofs, such as Dafny~\cite{leino2010dafny} or Lean~\cite{moura2021lean}, to model and verify these propositions. Future research should focus on refining proposition modeling and verification by prompting LLMs for code generation~\cite{mugnier2024laurel}, improving integration with external theorem provers and assertion verifiers (e.g., Dafny and Lean), and scaling P$^2$oT to handle multi-modal data for advanced reasoning. Further, automating code generation, optimizing hybrid P$^2$oT/CoT/ToT models, incorporating self-verification and self-correction, and adopting P$^2$oT into domain-specific applications, including logical deduction and scientific discovery can significantly advance its capabilities~\cite{romera2024mathematical}. Figure~\ref{fig:pot-exp} demonstrates the use of the P$^2$oT framework to verify a basic mathematical proof that for any even integer $n$, the sum of the first $n$ positive integers is an even number. By decomposing the problem into distinct, verifiable propositions and using Dafny for formal verification, this example highlights the structured and rigorous approach of P$^2$oT in logical reasoning and verification.

\section{Conclusions}
In conclusion, the synthesis of connectionist and symbolic paradigms, particularly through the rise of LLM-empowered Autonomous Agents (LAAs), marks a pivotal evolution in the field of AI, especially the neuro-symbolic AI. This paper has highlighted the historical context and the ongoing convergence of symbolic reasoning and neural network-based methods, underscoring how LAAs leverage the text-based knowledge representation and generative capabilities of LLMs to achieve logical reasoning and decision-making. By contrasting LAAs with Knowledge Graphs (KGs), we have demonstrated the unique advantages of LAAs in mimicking human-like reasoning processes, scaling effectively with large datasets, and leveraging in-context learning without extensive re-training. Promising directions such as neuro-vector-symbolic architectures and program-proof-of-thoughts (P$^2$oT) prompting are on the horizon, potentially enhancing the agentic reasoning capabilities of AI further. These insights not only encapsulate the transformative potential of current AI technologies but also provide a clear trajectory for future research, fostering a deeper understanding and more advanced applications of neuro-symbolic AI.  

\bibliographystyle{unsrt}
\bibliography{sn-bibliography}

\begin{thebibliography}{10}

\bibitem{lecun2015deep}
Yann LeCun, Yoshua Bengio, and Geoffrey Hinton.
\newblock Deep learning.
\newblock {\em Nature}, 521(7553):436--444, 2015.

\bibitem{newell2007computer}
Allen Newell and Herbert~A Simon.
\newblock Computer science as empirical inquiry: Symbols and search.
\newblock In {\em ACM Turing award lectures}, page 1975. 2007.

\bibitem{krizhevsky2012imagenet}
Alex Krizhevsky, Ilya Sutskever, and Geoffrey~E Hinton.
\newblock Imagenet classification with deep convolutional neural networks.
\newblock {\em Advances in neural information processing systems}, 25, 2012.

\bibitem{shortliffe2012computer}
Edward Shortliffe.
\newblock {\em Computer-based medical consultations: MYCIN}, volume~2.
\newblock Elsevier, 2012.

\bibitem{brown2020language}
Tom Brown, Benjamin Mann, Nick Ryder, Melanie Subbiah, Jared~D Kaplan, Prafulla
  Dhariwal, Arvind Neelakantan, Pranav Shyam, Girish Sastry, Amanda Askell,
  et~al.
\newblock Language models are few-shot learners.
\newblock {\em Advances in neural information processing systems},
  33:1877--1901, 2020.

\bibitem{radford2019language}
Alec Radford, Jeffrey Wu, Rewon Child, David Luan, Dario Amodei, Ilya
  Sutskever, et~al.
\newblock Language models are unsupervised multitask learners.
\newblock {\em OpenAI blog}, 1(8):9, 2019.

\bibitem{bubeck2023sparks}
S{\'e}bastien Bubeck, Varun Chandrasekaran, Ronen Eldan, Johannes Gehrke, Eric
  Horvitz, Ece Kamar, Peter Lee, Yin~Tat Lee, Yuanzhi Li, Scott Lundberg,
  et~al.
\newblock Sparks of artificial general intelligence: Early experiments with
  gpt-4.
\newblock {\em arXiv preprint arXiv:2303.12712}, 2023.

\bibitem{xiong2023natural}
Haoyi Xiong, Jiang Bian, Sijia Yang, Xiaofei Zhang, Linghe Kong, and Daqing
  Zhang.
\newblock Natural language based context modeling and reasoning with llms: A
  tutorial.
\newblock {\em arXiv preprint arXiv:2309.15074}, 2023.

\bibitem{evans2003two}
Jonathan St~BT Evans.
\newblock In two minds: dual-process accounts of reasoning.
\newblock {\em Trends in cognitive sciences}, 7(10):454--459, 2003.

\bibitem{bengio2020deep}
Yoshua Bengio.
\newblock Deep learning for system 2 processing.
\newblock {\em AAAI 2020}, 2020.

\bibitem{ji2021survey}
Shaoxiong Ji, Shirui Pan, Erik Cambria, Pekka Marttinen, and S~Yu Philip.
\newblock A survey on knowledge graphs: Representation, acquisition, and
  applications.
\newblock {\em IEEE Transactions on Neural Networks and Learning Systems},
  33(2):494--514, 2021.

\bibitem{webb2023emergent}
Taylor Webb, Keith~J Holyoak, and Hongjing Lu.
\newblock Emergent analogical reasoning in large language models.
\newblock {\em Nature Human Behaviour}, 7(9):1526--1541, 2023.

\bibitem{strachan2024testing}
James~WA Strachan, Dalila Albergo, Giulia Borghini, Oriana Pansardi, Eugenio
  Scaliti, Saurabh Gupta, Krati Saxena, Alessandro Rufo, Stefano Panzeri, Guido
  Manzi, et~al.
\newblock Testing theory of mind in large language models and humans.
\newblock {\em Nature Human Behaviour}, pages 1--11, 2024.

\bibitem{garcez2023neurosymbolic}
Artur~d’Avila Garcez and Luis~C Lamb.
\newblock Neurosymbolic ai: The 3 rd wave.
\newblock {\em Artificial Intelligence Review}, 56(11):12387--12406, 2023.

\bibitem{rosenblatt1958perceptron}
Frank Rosenblatt.
\newblock The perceptron: a probabilistic model for information storage and
  organization in the brain.
\newblock {\em Psychological review}, 65(6):386, 1958.

\bibitem{rumelhart1986learning}
David~E Rumelhart, Geoffrey~E Hinton, and Ronald~J Williams.
\newblock Learning representations by back-propagating errors.
\newblock {\em Nature}, 323(6088):533--536, 1986.

\bibitem{newell1956logic}
Allen Newell and Herbert Simon.
\newblock The logic theory machine--a complex information processing system.
\newblock {\em IRE Transactions on information theory}, 2(3):61--79, 1956.

\bibitem{buchanan1978dendral}
Bruce~G Buchanan and Edward~A Feigenbaum.
\newblock Dendral and meta-dendral: Their applications dimension.
\newblock {\em Artificial intelligence}, 11(1-2):5--24, 1978.

\bibitem{goel1989integration}
Ashok~Kumar Goel.
\newblock {\em Integration of case-based reasoning and model-based reasoning
  for adaptive design problem-solving}.
\newblock The Ohio State University, 1989.

\bibitem{lipton2018mythos}
Zachary~C Lipton.
\newblock The mythos of model interpretability: In machine learning, the
  concept of interpretability is both important and slippery.
\newblock {\em Queue}, 16(3):31--57, 2018.

\bibitem{feigenbaum1977art}
Edward~A Feigenbaum et~al.
\newblock The art of artificial intelligence: Themes and case studies of
  knowledge engineering.
\newblock 1977.

\bibitem{elkan1993building}
Charles Elkan and Russell Greiner.
\newblock Building large knowledge-based systems: Representation and inference
  in the cyc project: Db lenat and rv guha, 1993.

\bibitem{bengio2021deep}
Yoshua Bengio, Yann Lecun, and Geoffrey Hinton.
\newblock Deep learning for ai.
\newblock {\em Communications of the ACM}, 64(7):58--65, 2021.

\bibitem{garcez2008neural}
Artur~SD'Avila Garcez, Luis~C Lamb, and Dov~M Gabbay.
\newblock {\em Neural-symbolic cognitive reasoning}.
\newblock Springer Science \& Business Media, 2008.

\bibitem{devlin2018bert}
Jacob Devlin, Ming-Wei Chang, Kenton Lee, and Kristina Toutanova.
\newblock Bert: Pre-training of deep bidirectional transformers for language
  understanding.
\newblock {\em arXiv preprint arXiv:1810.04805}, 2018.

\bibitem{silver2017mastering}
David Silver, Julian Schrittwieser, Karen Simonyan, Ioannis Antonoglou, Aja
  Huang, Arthur Guez, Thomas Hubert, Lucas Baker, Matthew Lai, Adrian Bolton,
  et~al.
\newblock Mastering the game of go without human knowledge.
\newblock {\em Nature}, 550(7676):354--359, 2017.

\bibitem{W3C1999}
{World Wide Web Consortium (W3C)}.
\newblock Resource description framework (rdf) model and syntax specification,
  1999.
\newblock Retrieved from
  \url{https://www.w3.org/TR/1999/REC-rdf-syntax-19990222/}.

\bibitem{berners2023semantic}
Tim Berners-Lee, James Hendler, and Ora Lassila.
\newblock The semantic web: A new form of web content that is meaningful to
  computers will unleash a revolution of new possibilities.
\newblock In {\em Linking the World’s Information: Essays on Tim
  Berners-Lee’s Invention of the World Wide Web}, pages 91--103. 2023.

\bibitem{antoniou2004semantic}
Grigoris Antoniou and Frank Van~Harmelen.
\newblock {\em A semantic web primer}.
\newblock MIT press, 2004.

\bibitem{gruber1993translation}
Thomas~R Gruber.
\newblock A translation approach to portable ontology specifications.
\newblock {\em Knowledge Acquisition}, 5(2):199--220, 1993.

\bibitem{kok2005learning}
Stanley Kok and Pedro Domingos.
\newblock Learning the structure of markov logic networks.
\newblock In {\em Proceedings of the 22nd International Conference on Machine
  Learning}, pages 441--448, 2005.

\bibitem{nickel2015review}
Maximilian Nickel, Kevin Murphy, Volker Tresp, and Evgeniy Gabrilovich.
\newblock A review of relational machine learning for knowledge graphs.
\newblock {\em Proceedings of the IEEE}, 104(1):11--33, 2015.

\bibitem{kipf2016semi}
Thomas~N Kipf and Max Welling.
\newblock Semi-supervised classification with graph convolutional networks.
\newblock In {\em International Conference on Learning Representations}, 2022.

\bibitem{wang2017knowledge}
Quan Wang, Zhendong Mao, Bin Wang, and Li~Guo.
\newblock Knowledge graph embedding: A survey of approaches and applications.
\newblock {\em IEEE Transactions on Knowledge and Data Engineering},
  29(12):2724--2743, 2017.

\bibitem{hochreiter1997long}
Sepp Hochreiter and J{\"u}rgen Schmidhuber.
\newblock Long short-term memory.
\newblock {\em Neural computation}, 9(8):1735--1780, 1997.

\bibitem{vaswani2017attention}
Ashish Vaswani, Noam Shazeer, Niki Parmar, Jakob Uszkoreit, Llion Jones,
  Aidan~N Gomez, {\L}ukasz Kaiser, and Illia Polosukhin.
\newblock Attention is all you need.
\newblock {\em Advances in neural information processing systems}, 30, 2017.

\bibitem{raffel2020exploring}
Colin Raffel, Noam Shazeer, Adam Roberts, Katherine Lee, Sharan Narang, Michael
  Matena, Yanqi Zhou, Wei Li, and Peter~J Liu.
\newblock Exploring the limits of transfer learning with a unified text-to-text
  transformer.
\newblock {\em Journal of machine learning research}, 21(140):1--67, 2020.

\bibitem{achiam2023gpt}
Josh Achiam, Steven Adler, Sandhini Agarwal, Lama Ahmad, Ilge Akkaya,
  Florencia~Leoni Aleman, Diogo Almeida, Janko Altenschmidt, Sam Altman,
  Shyamal Anadkat, et~al.
\newblock Gpt-4 technical report.
\newblock {\em arXiv preprint arXiv:2303.08774}, 2023.

\bibitem{team2023gemini}
Gemini Team, Rohan Anil, Sebastian Borgeaud, Yonghui Wu, Jean-Baptiste Alayrac,
  Jiahui Yu, Radu Soricut, Johan Schalkwyk, Andrew~M Dai, Anja Hauth, et~al.
\newblock Gemini: a family of highly capable multimodal models.
\newblock {\em arXiv preprint arXiv:2312.11805}, 2023.

\bibitem{chowdhery2023palm}
Aakanksha Chowdhery, Sharan Narang, Jacob Devlin, Maarten Bosma, Gaurav Mishra,
  Adam Roberts, Paul Barham, Hyung~Won Chung, Charles Sutton, Sebastian
  Gehrmann, et~al.
\newblock Palm: Scaling language modeling with pathways.
\newblock {\em Journal of Machine Learning Research}, 24(240):1--113, 2023.

\bibitem{abdin2024phi}
Marah Abdin, Sam~Ade Jacobs, Ammar~Ahmad Awan, Jyoti Aneja, Ahmed Awadallah,
  Hany Awadalla, Nguyen Bach, Amit Bahree, Arash Bakhtiari, Harkirat Behl,
  et~al.
\newblock Phi-3 technical report: A highly capable language model locally on
  your phone.
\newblock {\em arXiv preprint arXiv:2404.14219}, 2024.

\bibitem{touvron2023llama}
Hugo Touvron, Thibaut Lavril, Gautier Izacard, Xavier Martinet, Marie-Anne
  Lachaux, Timoth{\'e}e Lacroix, Baptiste Rozi{\`e}re, Naman Goyal, Eric
  Hambro, Faisal Azhar, et~al.
\newblock Llama: Open and efficient foundation language models.
\newblock {\em arXiv preprint arXiv:2302.13971}, 2023.

\bibitem{liu2023agentbench}
Xiao Liu, Hao Yu, Hanchen Zhang, Yifan Xu, Xuanyu Lei, Hanyu Lai, Yu~Gu,
  Hangliang Ding, Kaiwen Men, Kejuan Yang, et~al.
\newblock Agentbench: Evaluating llms as agents.
\newblock {\em arXiv preprint arXiv:2308.03688}, 2023.

\bibitem{ding2023parameter}
Ning Ding, Yujia Qin, Guang Yang, Fuchao Wei, Zonghan Yang, Yusheng Su,
  Shengding Hu, Yulin Chen, Chi-Min Chan, Weize Chen, et~al.
\newblock Parameter-efficient fine-tuning of large-scale pre-trained language
  models.
\newblock {\em Nature Machine Intelligence}, 5(3):220--235, 2023.

\bibitem{ouyang2022training}
Long Ouyang, Jeffrey Wu, Xu~Jiang, Diogo Almeida, Carroll Wainwright, Pamela
  Mishkin, Chong Zhang, Sandhini Agarwal, Katarina Slama, Alex Ray, et~al.
\newblock Training language models to follow instructions with human feedback.
\newblock {\em Advances in neural information processing systems},
  35:27730--27744, 2022.

\bibitem{schaeffer2024emergent}
Rylan Schaeffer, Brando Miranda, and Sanmi Koyejo.
\newblock Are emergent abilities of large language models a mirage?
\newblock {\em Advances in Neural Information Processing Systems}, 36, 2024.

\bibitem{wei2022chain}
Jason Wei, Xuezhi Wang, Dale Schuurmans, Maarten Bosma, Fei Xia, Ed~Chi, Quoc~V
  Le, Denny Zhou, et~al.
\newblock Chain-of-thought prompting elicits reasoning in large language
  models.
\newblock {\em Advances in neural information processing systems},
  35:24824--24837, 2022.

\bibitem{li2023halueval}
Junyi Li, Xiaoxue Cheng, Wayne~Xin Zhao, Jian-Yun Nie, and Ji-Rong Wen.
\newblock Halueval: A large-scale hallucination evaluation benchmark for large
  language models.
\newblock In {\em Proceedings of the 2023 Conference on Empirical Methods in
  Natural Language Processing}, pages 6449--6464, 2023.

\bibitem{maes1993modeling}
Pattie Maes.
\newblock Modeling adaptive autonomous agents.
\newblock {\em Artificial life}, 1(1\_2):135--162, 1993.

\bibitem{albrecht2018autonomous}
Stefano~V Albrecht and Peter Stone.
\newblock Autonomous agents modelling other agents: A comprehensive survey and
  open problems.
\newblock {\em Artificial Intelligence}, 258:66--95, 2018.

\bibitem{arkin1998behavior}
Ronald~C Arkin.
\newblock {\em Behavior-based robotics}.
\newblock MIT press, 1998.

\bibitem{rizk2019cooperative}
Yara Rizk, Mariette Awad, and Edward~W Tunstel.
\newblock Cooperative heterogeneous multi-robot systems: A survey.
\newblock {\em ACM Computing Surveys (CSUR)}, 52(2):1--31, 2019.

\bibitem{koller2009probabilistic}
Daphne Koller and Nir Friedman.
\newblock {\em Probabilistic graphical models: principles and techniques}.
\newblock MIT press, 2009.

\bibitem{sutton2018reinforcement}
Richard~S Sutton and Andrew~G Barto.
\newblock {\em Reinforcement learning: An introduction}.
\newblock MIT press, 2018.

\bibitem{wang2024survey}
Lei Wang, Chen Ma, Xueyang Feng, Zeyu Zhang, Hao Yang, Jingsen Zhang, Zhiyuan
  Chen, Jiakai Tang, Xu~Chen, Yankai Lin, et~al.
\newblock A survey on large language model based autonomous agents.
\newblock {\em Frontiers of Computer Science}, 18(6):186345, 2024.

\bibitem{xi2023rise}
Zhiheng Xi, Wenxiang Chen, Xin Guo, Wei He, Yiwen Ding, Boyang Hong, Ming
  Zhang, Junzhe Wang, Senjie Jin, Enyu Zhou, et~al.
\newblock The rise and potential of large language model based agents: A
  survey.
\newblock {\em arXiv preprint arXiv:2309.07864}, 2023.

\bibitem{wu2023autogen}
Qingyun Wu, Gagan Bansal, Jieyu Zhang, Yiran Wu, Shaokun Zhang, Erkang Zhu,
  Beibin Li, Li~Jiang, Xiaoyun Zhang, and Chi Wang.
\newblock Autogen: Enabling next-gen llm applications via multi-agent
  conversation framework.
\newblock {\em arXiv preprint arXiv:2308.08155}, 2023.

\bibitem{LangChain2022}
Oguzhan Topsakal and Tahir~Cetin Akinci.
\newblock Creating large language model applications utilizing langchain: A
  primer on developing llm apps fast.
\newblock In {\em International Conference on Applied Engineering and Natural
  Sciences}, volume~1, pages 1050--1056, 2023.

\bibitem{Liu_LlamaIndex_2022}
Jerry Liu.
\newblock {LlamaIndex}, 11 2022.

\bibitem{song2023llm}
Chan~Hee Song, Jiaman Wu, Clayton Washington, Brian~M Sadler, Wei-Lun Chao, and
  Yu~Su.
\newblock Llm-planner: Few-shot grounded planning for embodied agents with
  large language models.
\newblock In {\em Proceedings of the IEEE/CVF International Conference on
  Computer Vision}, pages 2998--3009, 2023.

\bibitem{lin2024swiftsage}
Bill~Yuchen Lin, Yicheng Fu, Karina Yang, Faeze Brahman, Shiyu Huang, Chandra
  Bhagavatula, Prithviraj Ammanabrolu, Yejin Choi, and Xiang Ren.
\newblock Swiftsage: A generative agent with fast and slow thinking for complex
  interactive tasks.
\newblock {\em Advances in Neural Information Processing Systems}, 36, 2024.

\bibitem{johnson2019billion}
Jeff Johnson, Matthijs Douze, and Herv{\'e} J{\'e}gou.
\newblock Billion-scale similarity search with gpus.
\newblock {\em IEEE Transactions on Big Data}, 7(3):535--547, 2019.

\bibitem{guu2020retrieval}
Kelvin Guu, Kenton Lee, Zora Tung, Panupong Pasupat, and Mingwei Chang.
\newblock Retrieval augmented language model pre-training.
\newblock In {\em International conference on machine learning}, pages
  3929--3938. PMLR, 2020.

\bibitem{reedgeneralist}
Scott Reed, Konrad Zolna, Emilio Parisotto, Sergio~G{\'o}mez Colmenarejo,
  Alexander Novikov, Gabriel Barth-maron, Mai Gim{\'e}nez, Yury Sulsky, Jackie
  Kay, Jost~Tobias Springenberg, et~al.
\newblock A generalist agent.
\newblock {\em Transactions on Machine Learning Research}.

\bibitem{shen2024hugginggpt}
Yongliang Shen, Kaitao Song, Xu~Tan, Dongsheng Li, Weiming Lu, and Yueting
  Zhuang.
\newblock Hugginggpt: Solving ai tasks with chatgpt and its friends in hugging
  face.
\newblock {\em Advances in Neural Information Processing Systems}, 36, 2024.

\bibitem{gao2018neural}
Jianfeng Gao, Michel Galley, and Lihong Li.
\newblock Neural approaches to conversational ai.
\newblock In {\em The 41st international ACM SIGIR conference on research \&
  development in information retrieval}, pages 1371--1374, 2018.

\bibitem{yao2022react}
Shunyu Yao, Jeffrey Zhao, Dian Yu, Nan Du, Izhak Shafran, Karthik Narasimhan,
  and Yuan Cao.
\newblock React: Synergizing reasoning and acting in language models.
\newblock {\em arXiv preprint arXiv:2210.03629}, 2022.

\bibitem{karpas2022mrkl}
Ehud Karpas, Omri Abend, Yonatan Belinkov, Barak Lenz, Opher Lieber, Nir
  Ratner, Yoav Shoham, Hofit Bata, Yoav Levine, Kevin Leyton-Brown, et~al.
\newblock Mrkl systems: A modular, neuro-symbolic architecture that combines
  large language models, external knowledge sources and discrete reasoning.
\newblock {\em arXiv preprint arXiv:2205.00445}, 2022.

\bibitem{garcez2015neural}
Artur~d'Avila Garcez, Tarek~R Besold, Luc De~Raedt, Peter F{\"o}ldiak, Pascal
  Hitzler, Thomas Icard, Kai-Uwe K{\"u}hnberger, Luis~C Lamb, Risto
  Miikkulainen, and Daniel~L Silver.
\newblock Neural-symbolic learning and reasoning: contributions and challenges.
\newblock In {\em 2015 AAAI Spring Symposium Series}, 2015.

\bibitem{brachman2004knowledge}
Ronald Brachman and Hector Levesque.
\newblock {\em Knowledge representation and reasoning}.
\newblock Morgan Kaufmann, 2004.

\bibitem{yao2024tree}
Shunyu Yao, Dian Yu, Jeffrey Zhao, Izhak Shafran, Tom Griffiths, Yuan Cao, and
  Karthik Narasimhan.
\newblock Tree of thoughts: Deliberate problem solving with large language
  models.
\newblock {\em Advances in Neural Information Processing Systems}, 36, 2024.

\bibitem{romera2024mathematical}
Bernardino Romera-Paredes, Mohammadamin Barekatain, Alexander Novikov, Matej
  Balog, M~Pawan Kumar, Emilien Dupont, Francisco~JR Ruiz, Jordan~S Ellenberg,
  Pengming Wang, Omar Fawzi, et~al.
\newblock Mathematical discoveries from program search with large language
  models.
\newblock {\em Nature}, 625(7995):468--475, 2024.

\bibitem{min2022rethinking}
Sewon Min, Xinxi Lyu, Ari Holtzman, Mikel Artetxe, Mike Lewis, Hannaneh
  Hajishirzi, and Luke Zettlemoyer.
\newblock Rethinking the role of demonstrations: What makes in-context learning
  work?
\newblock {\em arXiv preprint arXiv:2202.12837}, 2022.

\bibitem{delong2023neurosymbolic}
Lauren~Nicole DeLong, Ramon~Fern{\'a}ndez Mir, and Jacques~D Fleuriot.
\newblock Neurosymbolic ai for reasoning over knowledge graphs: A survey.
\newblock {\em arXiv preprint arXiv:2302.07200}, 2023.

\bibitem{peng2023knowledge}
Ciyuan Peng, Feng Xia, Mehdi Naseriparsa, and Francesco Osborne.
\newblock Knowledge graphs: Opportunities and challenges.
\newblock {\em Artificial Intelligence Review}, 56(11):13071--13102, 2023.

\bibitem{xie2021explanation}
Sang~Michael Xie, Aditi Raghunathan, Percy Liang, and Tengyu Ma.
\newblock An explanation of in-context learning as implicit bayesian inference.
\newblock {\em arXiv preprint arXiv:2111.02080}, 2021.

\bibitem{askari2024self}
Arian Askari, Roxana Petcu, Chuan Meng, Mohammad Aliannejadi, Amin Abolghasemi,
  Evangelos Kanoulas, and Suzan Verberne.
\newblock Self-seeding and multi-intent self-instructing llms for generating
  intent-aware information-seeking dialogs.
\newblock {\em arXiv preprint arXiv:2402.11633}, 2024.

\bibitem{muennighoff2024generative}
Niklas Muennighoff, Hongjin Su, Liang Wang, Nan Yang, Furu Wei, Tao Yu,
  Amanpreet Singh, and Douwe Kiela.
\newblock Generative representational instruction tuning.
\newblock {\em arXiv preprint arXiv:2402.09906}, 2024.

\bibitem{velivckovic2021neural}
Petar Veli{\v{c}}kovi{\'c} and Charles Blundell.
\newblock Neural algorithmic reasoning.
\newblock {\em Patterns}, 2(7), 2021.

\bibitem{kanerva2009hyperdimensional}
Pentti Kanerva.
\newblock Hyperdimensional computing: An introduction to computing in
  distributed representation with high-dimensional random vectors.
\newblock {\em Cognitive computation}, 1:139--159, 2009.

\bibitem{hersche2023neuro}
Michael Hersche, Mustafa Zeqiri, Luca Benini, Abu Sebastian, and Abbas Rahimi.
\newblock A neuro-vector-symbolic architecture for solving raven’s
  progressive matrices.
\newblock {\em Nature Machine Intelligence}, 5(4):363--375, 2023.

\bibitem{leino2010dafny}
K~Rustan~M Leino.
\newblock Dafny: An automatic program verifier for functional correctness.
\newblock In {\em International conference on logic for programming artificial
  intelligence and reasoning}, pages 348--370. Springer, 2010.

\bibitem{moura2021lean}
Leonardo~de Moura and Sebastian Ullrich.
\newblock The lean 4 theorem prover and programming language.
\newblock In {\em Automated Deduction--CADE 28: 28th International Conference
  on Automated Deduction, Virtual Event, July 12--15, 2021, Proceedings 28},
  pages 625--635. Springer, 2021.

\bibitem{mugnier2024laurel}
Eric Mugnier, Emmanuel~Anaya Gonzalez, Ranjit Jhala, Nadia Polikarpova, and
  Yuanyuan Zhou.
\newblock Laurel: Generating dafny assertions using large language models.
\newblock {\em arXiv preprint arXiv:2405.16792}, 2024.

\end{thebibliography}

\end{document}